\documentclass[10pt,twocolumn,letterpaper]{article}

\usepackage[pagenumbers]{cvpr} 

\usepackage{graphicx}
\usepackage{amsmath}
\usepackage{amssymb}
\usepackage{booktabs}

\usepackage[linesnumbered,ruled]{algorithm2e}
\usepackage[dvipsnames]{xcolor}
\usepackage{multirow}
\usepackage{diagbox}

\usepackage[pagebackref,breaklinks,colorlinks]{hyperref}

\usepackage[capitalize]{cleveref}
\crefname{section}{Sec.}{Secs.}
\Crefname{section}{Section}{Sections}
\Crefname{table}{Table}{Tables}
\crefname{table}{Tab.}{Tabs.}

\def\cvprPaperID{1485} 
\def\confName{CVPR}
\def\confYear{2022}

\begin{document}

\title{Cross-Modal Transferable Adversarial Attacks from Images to Videos}

\author{
    Zhipeng Wei \\
    Fudan University \\
    {\tt\small zpwei21@m.fudan.edu.cn} 
    \and
    Jingjing Chen\\
    Fudan University \\
    {\tt\small chenjingjing@fudan.edu.cn} 
    \and
    Zuxuan Wu\\
    Fudan University \\
    {\tt\small zxwu@fudan.edu.cn} 
    \and
    Yu-Gang Jiang \\
    Fudan University \\
    {\tt\small ygj@fudan.edu.cn} 
}
\maketitle

\begin{abstract}
Recent studies have shown that adversarial examples hand-crafted on one white-box model can be used to attack other black-box models. 
Such cross-model transferability makes it feasible to perform black-box attacks, which has raised security concerns for real-world DNNs applications.
Nevertheless, existing works mostly focus on investigating the adversarial transferability across different deep models that share the same modality of input data. The cross-modal transferability of adversarial perturbation has never been explored.
This paper investigates the transferability of adversarial perturbation across different modalities, i.e., leveraging adversarial perturbation generated on white-box image models to attack black-box video models. Specifically, motivated by the observation that the low-level feature space between images and video frames are similar, we propose a simple yet effective cross-modal attack method, named as Image To Video (I2V) attack. I2V generates adversarial frames by minimizing the cosine similarity between features of pre-trained image models from adversarial and benign examples, then combines the generated adversarial frames to perform black-box attacks on video recognition models. 
Extensive experiments demonstrate that I2V can achieve high attack success rates on different black-box video recognition models. On Kinetics-400 and UCF-101, I2V achieves an average attack success rate of 77.88\% and 65.68\%, respectively, which sheds light on the feasibility of cross-modal adversarial attacks.
\end{abstract}

\section{Introduction}
Deep learning has achieved significant progress in a series of computer vision tasks, such as image recognition \cite{he2016deep}, object detection \cite{ren2015faster}, action recognition \cite{carreira2017quo}. 
However, recent studies have shown that deep neural networks (DNNs) are highly vulnerable to adversarial examples \cite{szegedy2013intriguing,goodfellow2014explaining}, which are generated by adding small human-imperceptible perturbations that can lead to wrong predictions. 
The existence of adversarial examples has posed serious security threats for the application of DNNs in security-critical scenarios, such as autonomous driving \cite{sharif2016accessorize}, face recognition \cite{eykholt2018robust}, video analysis \cite{wei2020heuristic}, etc. As a result, adversarial examples have attracted numerous research attentions in recent years.

\begin{figure*}[t]
\centering
\includegraphics[width=0.85\textwidth]{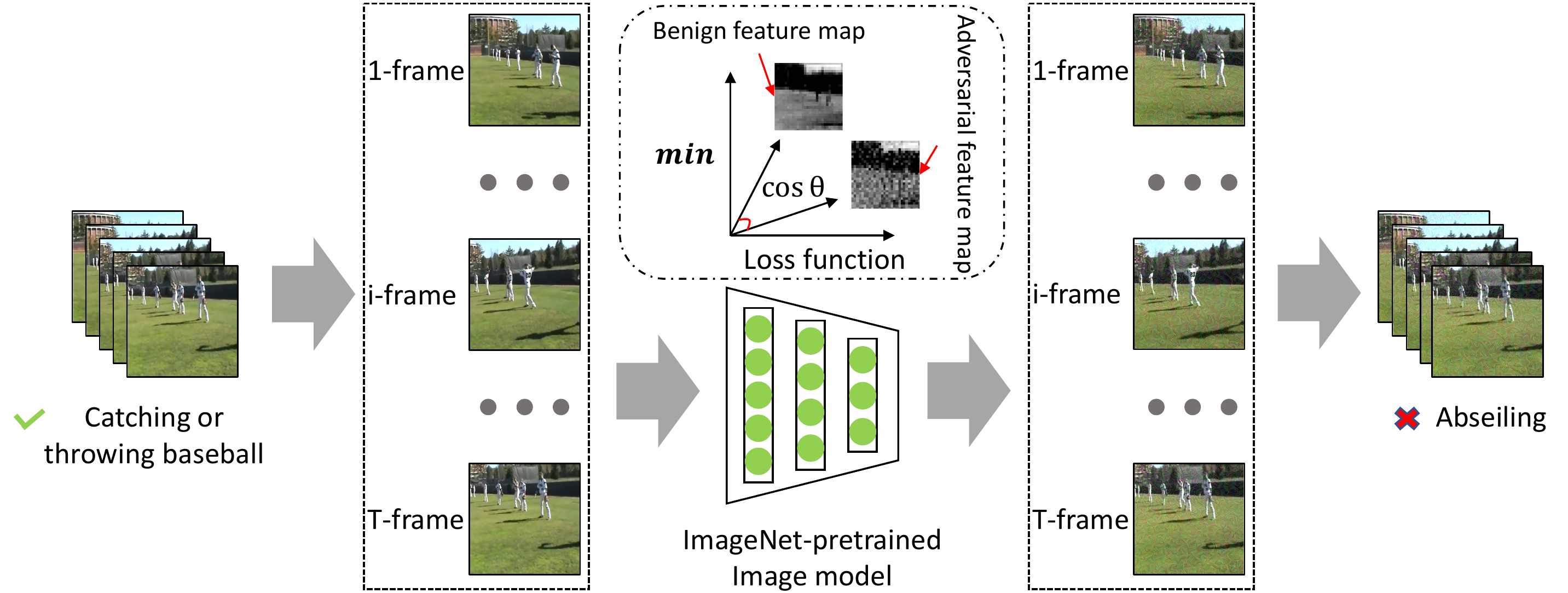}
\caption{Overview of the proposed I2V attack. Given a video clip with a true label of ``Catching or throwing baseball", where each frame is input into the ImageNet-pretrained image model separately. Then the image model generates adversarial frames by minimizing the cosine similarity between features from adversarial and benign examples. As the image model and the video model share similar feature space, the generated video adversarial example can fool the video recognition models, and be misclassified as ``Abseiling".}
\label{overview}
\end{figure*}

It has been demonstrated in recent works \cite{liu2016delving} that adversarial examples have the property of transferability, i.e., the adversarial example generated from one model can be used to attack other models. Such cross-model transferability makes it feasible to perform black-box attacks by leveraging adversarial examples handcrafted on white-box models. As a result, how to enhance the transferability of adversarial examples for efficient black-box attacks has attracted several research interests recently. 
These works either perform data augmentation \cite{xie2019improving, lin2019nesterov, dong2019evading}, optimize gradient calculations \cite{dong2018boosting, lin2019nesterov, wu2020skip}, or disrupt common properties among different models \cite{wu2020boosting}, to avoid the generated adversarial samples being overfitted to white-box models. 
Nevertheless, all of these works require the white-box models and target black-box models to be homomodal, which share the same modal of input data. Transferability between hetero-modal models has never been explored.

To bridge this gap, this paper investigates the cross-modal transferability of adversarial examples.
Specifically, we explore the adversarial transferability between image models and video models by performing transfer-based black-box attacks on video models with image models pre-trained on ImageNet only. This is an extremely challenging setting since there are no white-box video models for generating video adversarial samples. There are two major obstacles in transferring adversarial perturbations generated on images models to attack video models. First, in addition to the domain gap between image and video data, video data contain additional temporal information, which leads to differences in the learned features between image models and video models. The difference makes it difficult to transfer the adversarial perturbations from images to videos. 
Second, existing transfer-based attacks on homomodal models (e.g., image models) are not applicable to the cross-modal attack scenarios.  Unlike existing transfer-based attacks on images, where image labels are available for optimizing task-specific loss function (e.g., Cross Entropy loss) in the process of adversarial perturbations generation, in the cross-modal image to video attack, no labels are available for video frames.

To address the aforementioned challenges and perform black-box attacks on video models, we propose a simple yet effective cross-modal attack method, named Image To Video (I2V) attack. Despite there being a domain gap between image data and video data, we observed that the intermediate features between images models and video models are similar to a certain extent. 
This motivates us to perturb the intermediate features of the ImageNet-pretrained image models to craft adversarial video frames for attacking video recognition models. To this end, the proposed I2V optimizes the adversarial perturbations by minimizing the cosine similarity of intermediate features between benign frames and the generated adversarial frames. 
The minimization of the cosine similarity makes the features extracted from adversarial video frames orthogonal to that from benign frames. Consequently, it will cause the adversarial video features to move away from the benign video features due to the feature similarity between image and video models.
Figure \ref{overview} gives an overview of the proposed I2V attack method. I2V takes individual frames from a video clip as input to the image model and generates adversarial frames one by one. Then the generated adversarial frames are grouped into video adversarial examples according to the temporal information of the benign video clip. 
We briefly summarize our primary contributions as follows:
\begin{itemize}
\item We investigate the transferability of adversarial perturbations between image models and video models. Specially, we propose an I2V attack to boost the transferability of video adversarial examples generated from image models across different video recognition models. To the best of our knowledge, this is the first work on cross-modal transfer-based black-box attacks for video recognition models.
\item We provide insightful analysis on the correlations of feature maps between image and video models. Based on this observation, I2V optimizes adversarial frames on perturbing feature maps of image models to boost the transferability across different video recognition models.
\item We conduct empirical evaluations using six video recognition models trained with the Kinetic-400 dataset and UCF-101 dataset. Extensive experiments demonstrate that our proposed I2V helps to boost the transferability of video adversarial examples generated from image models.
\end{itemize}

\section{Related work}
\subsection{Transfer-based attacks on image models}
\label{related_image}
Prior works in generating adversarial examples with high transferability are based on white-box attacks, such as Fast Gradient Sign Method (FGSM) \cite{goodfellow2014explaining} and Basic Iterative Method (BIM) \cite{kurakin2016adversarial}. FGSM linearizes the loss function around the current parameters and performs a one-step update along with the gradient sign of the loss function with respect to inputs. BIM is the iterative version of FGSM and overfits the white-box model to generate stronger adversarial examples in attacking the white-box model.
To further improve the transferability of adversarial examples in attacking black-box models, several approaches are proposed recently. In general, there are three ways to improve transferability, which include data augmentation, gradient modification and common property disruption of discriminative features among different models. 
The main idea of data augmentation is to improve the generalization of adversarial examples and avoid overfitting the white-box model. For example, Diversity Input (DI) \cite{xie2019improving} attack conducts random resizing and padding to the input. Scale-invariant method (SIM) \cite{lin2019nesterov} applies the scale transformation to the input. Translation-invariant (TI) \cite{dong2019evading} attack performs horizontal and vertical shifts with a short distance to the input.
The second way modifies gradients used for updating adversarial perturbations. For example, Momentum Iterative (MI) \cite{dong2018boosting} attack integrates the momentum into the iterative process for stabilizing update directions. As an improved momentum method, Nesterov Accelerated Gradient (NAG) \cite{lin2019nesterov} can be also integrated into the BIM. Skip Gradient Method (SGM) \cite{wu2020skip} uses more gradients from the skip connections and emphasizes the gradients of shallow layers.
The main idea of the third way focuses on disrupting the common property of classification among different models. For example, Attention-guided Transfer Attack (ATA) \cite{wu2020boosting} prioritizes the corruption of critical features that are likely to be adopted by diverse architectures.
Other transfer-based attacks such as Dispersion Reduction (DR) \cite{lu2020enhancing}, Intermediate Level Attack (ILA) \cite{huang2019enhancing} have improved the transferability of adversarial examples in different tasks through perturbing feature maps.
In contrast, the proposed I2V attack implements a cross-modal transfer-based attack through a correlation between the spatial features encoded between the image model and the video model.

\subsection{Transfer-based attacks on video recognition models}
\label{related_video}
There is much less work about transfer-based attacks on video models compared with transfer-based attacks on image models.
Temporal Translation (TT) attack method \cite{wei2021boosting} optimizes the adversarial perturbations over a set of temporal translated video clips for avoiding overfitting to the white-box model being attacked. Although TT achieves better results than transfer-based image attack methods, it increases the computational cost. Different from it, the proposed I2V attack achieves better performance without trained video models and is easy to perform.

\subsection{Video recognition models}
\label{realted_model}
Video action recognition models have made significant progress in recent years. 
Previous studies \cite{yue2015beyond,jiang2017exploiting} adopt a 2D + 1D paradigm, where 2D CNNs are applied over per-frame input to extract features, followed by a 1D module (\eg, RNNs) that integrates per-frame features. Current studies use 3D CNNs to jointly capture the dynamic semantic of videos. For example, I3D \cite{carreira2017quo} leverages ImageNet architecture designs and their parameters to encode spatio-temporal features by inflating the 2D convolution kernels into 3D. Non-local (NL) \cite{wang2018non} network inserts a non-local operation into I3D for encoding long-range temporal dependencies between video frames. SlowFast \cite{feichtenhofer2019slowfast} contrasts the visual tempos along the temporal axis, which involves a slow pathway and a fast pathway to capture spatial semantics and motion at fine temporal resolution respectively.
Temporal Pyramid Network (TPN) \cite{yang2020temporal} capture action instances at various tempos through a feature hierarchy architecture. 
In this paper, we use six representative video action recognition models for experiments, including NL, SlowFast, TPN with 3D Resnet-50 and Resnet-101 as backbones.

\section{Methodology}
\subsection{Preliminary}
Given a video sample $x \in \mathcal{X} \subset \mathbf{R}^{T \times H \times W \times C}$ with the true label $y \in \mathcal{Y} = \{1, 2, ..., K\}$, where $T$, $H$, $W$, $C$ denote the number of frames, height, width and channels respectively. $K$ represents the number of classes. Let $g$ denote the ImageNet-pretrained image model (\eg, ResNet, VGG), $f$ denote the video recognition model. We use $f(x) : \mathcal{X} \to \mathcal{Y} $ to denote the prediction of the video recognition model for an input video. Thus, the proposed I2V attack aims to generate the adversarial example $x_{adv}=x + \delta $ by $g$, which can fool the video model $f$ into $f(x_{adv}) \neq y$ without knowledge about $f$, where $\delta$ denotes the adversarial perturbation. To ensure that the adversarial perturbation $\delta$ is imperceptible, we restrict it by $||\delta||_{p} \leq \epsilon$, where $||\cdot||_{p}$ denotes the $L_{p}$ norm, and $\epsilon$ is a constant of the norm constraint. We adopt $L_{\infty}$ norm and untargeted adversarial attacks, which are commonly used in \cite{xie2019improving,lin2019nesterov,dong2019evading,wu2020skip,wu2020boosting}.
In a white-box setting, the objective of untargeted adversarial attacks can be formulated as follows:
\begin{equation}
    \label{eq1}
        \mathop{\arg\max}_{\delta} J(f(x+\delta),y), \\
        s.t.\,\, ||\delta||_{\infty}\le\epsilon,
\end{equation}
where $J$ is the loss function (\eg, cross-entropy loss) of the video model $f$. 
However, in this paper, the adversary cannot access knowledge about $f$. The proposed I2V attack leverages adversarial examples generated from $g$ to attack $f$ in the black-box setting.

\subsection{Correlation analysis between image and video models} 
Before introducing the proposed method, we firstly give an empirical analysis of the correlation between image and video models. It has been demonstrated in the prior work \cite{jiang2019black} that utilizing ImageNet-pretrained image models to generate tentative perturbations assumes fewer queries for attacking black-box video recognition models. This basically suggests that \textit{the intermediate features between images models and video models may be similar to a certain extent}. Hence perturbing intermediate feature maps of image models could affect that of video models.
To verify this assumption, we analyze the intermediate features' similarity of both benign and adversarial frames between image and video models with cosine similarity. 

\begin{figure}[t]
\centering
\includegraphics[width=0.9\columnwidth]{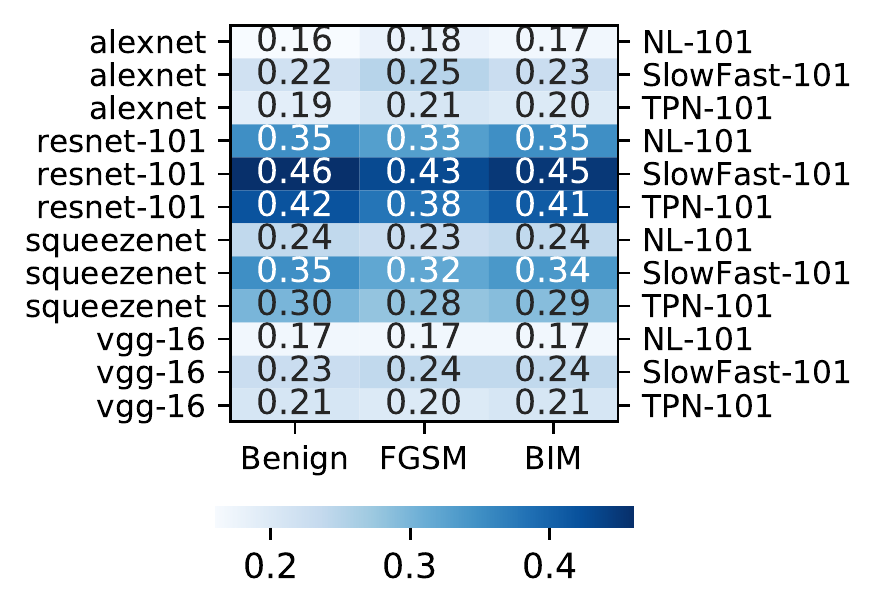}
\caption{Cosine similarity analysis of intermediate features between image models and video models on benign examples and adversarial examples. The cosine similarities are calculated on the averaged intermediate features extracted from 400 randomly selected videos in Kinetics-400. Adversarial video samples are generated by FGSM and BIM, using video models as white-box models. Darker color represents a higher cosine similarity. 
}
\label{fig_cos_sim_layer0}
\end{figure}

Figure~\ref{fig_cos_sim_layer0} shows the cosine similarity of intermediate features between image and video models. The intermediate features are extracted from 400 randomly selected videos in Kinetics-400 and then averaged for calculating cosine similarity. For all video models, the intermediate features are extracted from the first 3D-Resnet block, while for different image models, the features are extracted from different intermediate layers, which are summarized in Table~\ref{tab:attacked_layer} (marked in red color).
Here we choose different intermediate layers for different image models for the purpose of maximizing the similarity between image features and video features. From figure~\ref{fig_cos_sim_layer0}, it can be observed that, for both benign samples and adversarial samples, their intermediate layer features extracted from image models and video models are similar to a certain extent. This is mainly because the convolution operations in image models and video models are somehow similar. 
It is worthwhile to mention that the cosine similarities obtained from benign samples and adversarial samples are quite similar. This basically suggests that adversarial perturbations have little effect on the similarity of feature space between image and video models. Similar trends can be also observed when using other intermediate layers of video models (see Appendix).

\begin{figure}
  \centering
  \begin{subfigure}{.45\textwidth}
    \includegraphics[width=1.\linewidth]{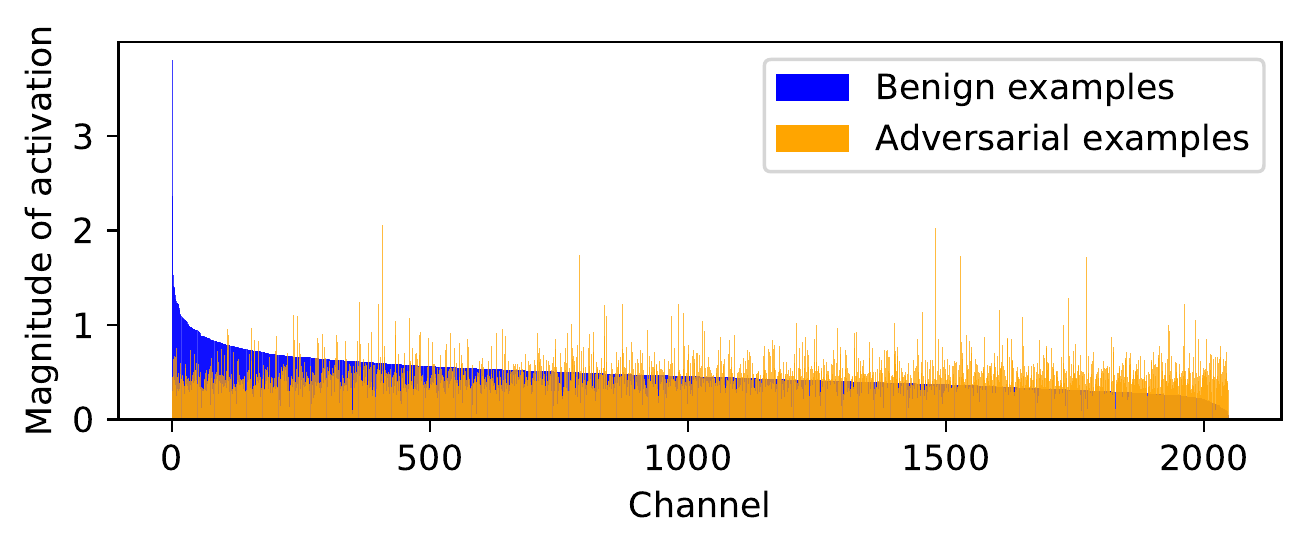}
    \caption{Resnet-101}
    \label{ss}
  \end{subfigure}
  \begin{subfigure}{.45\textwidth}
  \includegraphics[width=1.\linewidth]{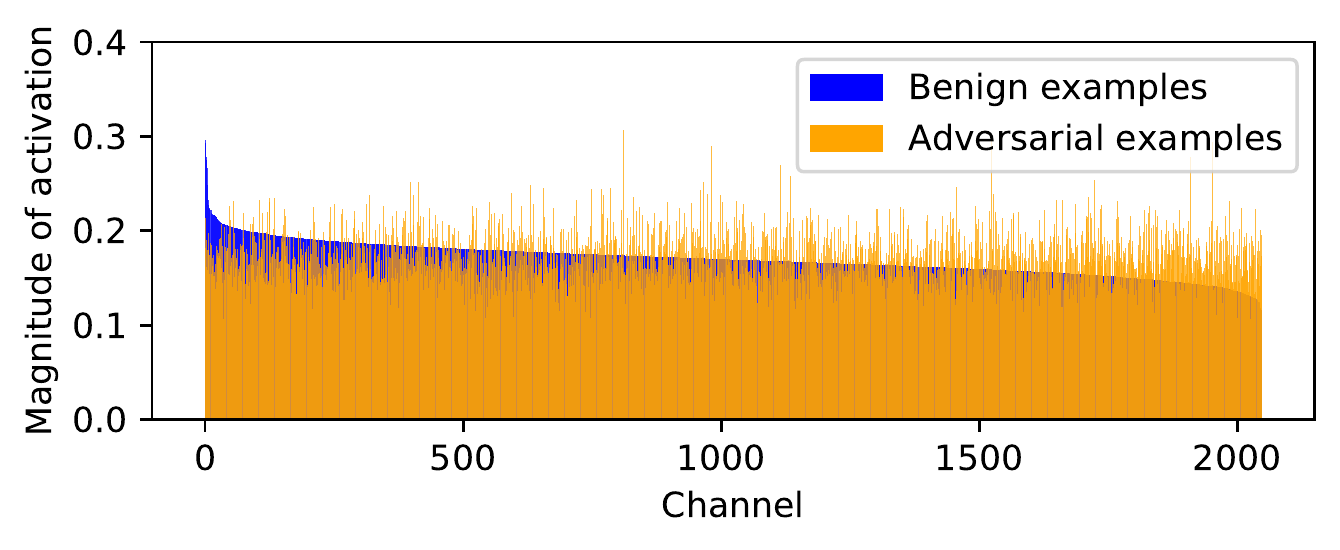}
    \caption{NL-101}
    \label{images/video_channel.pdf}
  \end{subfigure}
  \caption{The magnitudes (y-axis) of channel-wise activation at the penultimate layer (2048 channels at x-axis) for both image and video models. 
  The magnitude is calculated by the global average pooling at each channel. 
  In each plot, the channel-wise magnitudes are averaged from randomly selected 400 videos of Kinetics-400 and displayed separately for benign and adversarial examples (generated by BIM). The 2048 channels are sorted in descending order of magnitude for benign examples.}
  \label{fig_channel}
\end{figure}

To demonstrate that the adversarial perturbations on the feature maps are translatable between video and image models, we further compare the magnitude changes in the channel-wise activation of image and video models before and after adding the same adversarial perturbations to the video frames. The results are shown in Figure \ref{fig_channel}.
As can be seen, the adversarial examples generated on NL-101 perturb the channel-wise activation magnitude not only in NL-101 but also in Resnet-101.
Since each channel of features captures a specific pattern of the object and contributes differently to the final classification, the magnitude changes of image and video models are likely to result in wrong predictions, demonstrating possibilities in transferring adversarial perturbations between image and video models.

\subsection{Image To Video (I2V) Attack}
Based on the above observations, we propose the Image To Video (I2V) attack, which generates video adversarial examples from an ImageNet-pretrained image model, to boost the transferability of hetero-modal models and attack video models in the black-box setting.
By perturbing intermediate features of image models, I2V generates adversarial examples to disturb intermediate features of black-box video models with high probability. 
In particular, I2V optimizes the $i$-th adversarial frame by:
\begin{equation}
    \label{i2v}
        \mathop{\arg\min}_{\delta} CosSim(g_{l}(x^i+\delta),g_{l}(x^i)), \\
        s.t.\,\, ||\delta||_{\infty} \le \epsilon,
\end{equation}
where $g_{l}(x^i)$ denotes the intermediate feature map of the $l$-th layer with respect to $x^i$ in the image model, $x^{i} \in \mathbf{R}^{H \times W \times C}$ denotes the $i$-th frame of $x$, the function $CosSim$ calculates the cosine similarity between $g_{l}(x^i+\delta)$ and $g_{l}(x^i)$. 

In this way, the minimization of the cosine similarity makes it possible to optimize adversarial examples with features that are orthogonal to ones of benign examples. Consider that $g_{l}(x^i)$ is the output of the penultimate layer and let $W=(W_1, ..., W_y, ..., W_K)$ denote the weight of the classification layer, thus $W_y$ and $g_{l}(x^i)$ are highly aligned for making a true prediction. With minimizing $CosSim(g_{l}(x^i+\delta), g_{l}(x^i)) = \frac{g_{l}(x^i+\delta)^{T}g_{l}(x^i)}{||g_{l}(x^i+\delta)||\cdot ||g_{l}(x^i)||}$, we can get minimizing $g_{l}(x^i+\delta)^{T}g_{l}(x^i)$ if $g_{l}(x^i+\delta)$ and $g_{l}(x^i)$ have unit length. Due to the high alignment between $W_y$ and $g_{l}(x^i)$, the minimization of the cosine similarity induces that the value of $W_y \cdot g_{l}(x^i+\delta)$ decreases a lot to fool the image model $g$ into making error predictions. Based on the similarity of feature space between image and video models, the generated adversarial examples $x_{adv}=(x_{adv}^1, ... , x_{adv}^i, ..., x_{adv}^T)$ may fool video models with high probability by perturbing video intermediate features.

\begin{algorithm}[t]
    \SetKwInOut{Input}{Input}
    \SetKwInOut{Parameter}{Parameter}
    \SetKwInOut{Output}{Output}
    
    \Input{A video example $x$, image model $g$.}
    \Parameter{Perturbation budget $\epsilon$, iteration number $I$, step size $\alpha$, the number of layer $l$.}
    \Output{The adversarial example $x_{adv}$.}
    \For{$i=1$ \KwTo $T$}
        {$x^i$ = the $i$-th frame of $x$ \\ 
         $\delta_0^i = \mathbf{(\frac{0.01}{255})}^{H \times W \times C}$ \\
         \For{$j = 0$ \KwTo $I-1$}
         {Update $\delta_{j}^{i}$ by Adam optimizer:
         \, \, $\delta_{j+1}^{i} = ADAM(\delta_j^{i}, \alpha, CosSim(g_{l}(x^i+\delta^i_j),g_{l}(x^i)))$
         }
         Project $x^i_{adv}$ to the $\epsilon$-ball of $x^i$: \\
         \, \, $x_{adv}^{i} = clip_{x^i, \epsilon}(x^i + \delta_I^{i})$ 
        }
    \Return $x_{adv}=(x_{adv}^1, ..., x_{adv}^i, ..., x_{adv}^T)$
\caption{Image to Video (I2V) attack.}
\label{alg}
\end{algorithm}

Following \cite{wei2019sparse}, we initialize the adversarial perturbations $\delta$ with a small constant value $\frac{0.01}{255}$ and use the Adam optimizer \cite{kingma2014adam} to solve the Equation \ref{i2v} and updates $\delta^i_j$. Algorithm \ref{alg} illustrates the generation of adversarial examples of the proposed I2V attack. Where $I$ denotes the iteration number of Adam optimizer, $clip_{x^i, \epsilon}$ denote to project $x^i+\delta^i_I$ to the vicinity of $x^i$ for meeting $||\delta^i_I||_{\infty} \le \epsilon$. In the end, I2V attack combines all generated adversarial frames $x_{adv}^i$ into a video adversarial examples $x_{adv}$. 

\subsection{Attacking an ensemble of models}
MIFGSM \cite{dong2018boosting} shows that attacking an ensemble of models can boost the transferability of generated adversarial examples. When a generated example remains adversarial over an ensemble of models, it may transfer to attack other models. Based on this, we propose to use multiple ImageNet-pretrained image models to perform the I2V attack, named ENS-I2V, which optimizes $i$-th adversarial frame by:
\begin{equation}
    \label{ens_i2v}
        \mathop{\arg\min}_{\delta} \sum_{n=1}^{N} CosSim(g_{l}^n(x^i+\delta),g_{l}^n(x^i)), \\
        s.t.\,\, ||\delta||_{\infty} \le \epsilon,
\end{equation}
where $N$ is the number of used image models, $g_l^n(\cdot)$ returns the intermediate feature map of the $l$-th layer in the $n$-th image model. 
In this way, the intermediate features of the adversarial frames generated by ENS-I2V are orthogonal to the ensemble of features from benign examples, thus ENS-I2V allows the generation of highly transferable adversarial examples.

\section{Experiments}
\subsection{Experimental setting}
\textbf{Dataset.} 
We evaluate our approach using UCF-101 \cite{soomro2012ucf101} and Kinetics-400 \cite{carreira2017quo} datasets, which are widely used datasets for video recognition. 
UCF-101 consists of 13,320 videos from 101 actions. 
Kinetics-400 contains approximately 240,000 videos from 400 human actions.

\textbf{ImageNet-pretrained image models.}
We perform our proposed methods on four ImageNet-pretrained image models: Alexnet \cite{krizhevsky2014one}, Resnet-101 \cite{he2016deep}, Squeezenet 1.1 \cite{iandola2016squeezenet} and Vgg-16 \cite{simonyan2014very}. Where Squeezenet 1.1 has 2.4x less computation and slightly fewer parameters than SqueezeNet 1.0, without sacrificing accuracy. These four models are commonly used in the image classification.

\begin{table}[]
    \centering
    \begin{tabular}{l c c} \toprule
          Model & UCF-101 & Kinetics-400 \\ \midrule
          NL-50 & 81.26 & 75.17 \\ 
          NL-101 & 82.21 & 75.81 \\ 
          SlowFast-50 & 85.25 & 76.66 \\ 
          SlowFast-101 & 86.10 & 76.95 \\
          TPN-50 & 87.13& 78.90 \\ 
          TPN-101 & 90.28 & 79.70 \\ \bottomrule
    \end{tabular}
    \caption{Top-1 validation accuracy(\%) of video recognition models on UCF-101 and Kinetics-400.}
    \label{tab:acc}
\end{table}

\textbf{Video recognition models.}
Our proposed methods are evaluated on three different architectures of video recognition models: Non-local (NL) \cite{wang2018non}, SlowFast \cite{feichtenhofer2019slowfast}, TPN \cite{yang2020temporal}. NL, SlowFast and TPN use 3D Resnet-50/101 as the backbone.
We train these video models from scratch with Kinetics-400 and fine-tune them on UCF-101. 
For Kinetics-400, we skip every other frame from randomly selected 64 consecutive frames into constructing input clips. 
For UCF-101, we use 32 consecutive frames as input clips.
The spatial size of the input is $224 \times 224$. Table \ref{tab:acc} summarizes top-1 validation accuracy of these six models on UCF-101 and Kinetics-400.

\textbf{Attack setting.}
In our experiments, we use the Attack Success Rate (ASR) to evaluate the attack performance, which is the rate of adversarial examples that are successfully misclassified by the black-box video recognition model. Thus higher ASR means better adversarial transferability. If not specifically stated, average ASR (AASR) is the average ASR over all black-box video models.
Following \cite{dong2019evading, xie2019improving}, we randomly sample one video, which is correctly classified by all models, from each class to conduct our experiments, and set the norm constraint $\epsilon=16$. 

\subsection{Ablation study}
We first investigate the effects of step size $\alpha$, iteration number $I$ and different attacked layers $l$ of image models. The evaluations are conducted on video models trained on Kinetics-400.

\begin{figure}[t]
\centering
    \includegraphics[width=0.75\columnwidth]{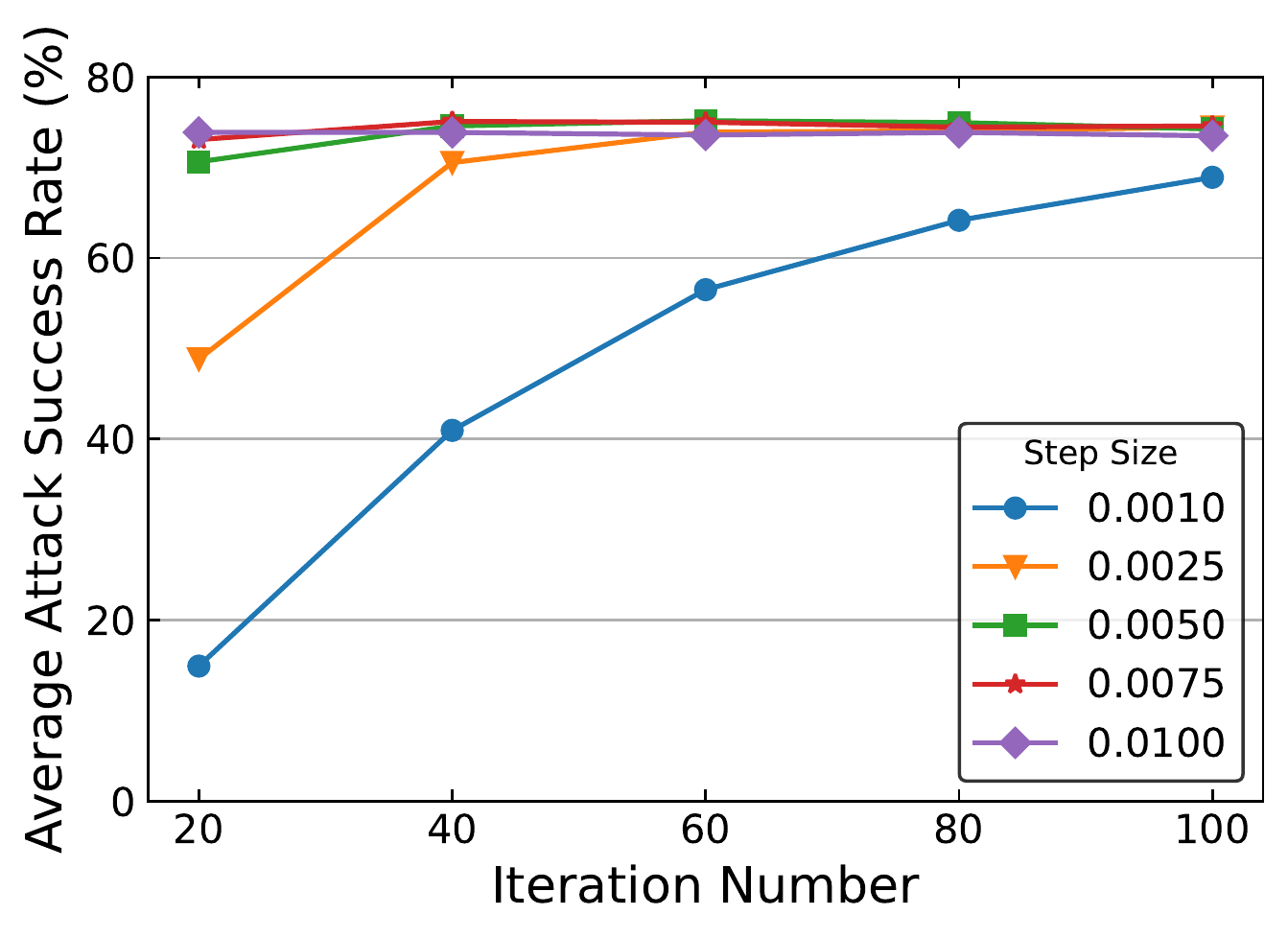}
\caption{AASR (\%) of the I2V attack with various step sizes and iteration numbers.}
\label{fig_abl_step_iter}
\end{figure}

\textbf{Step size and iteration number.}
Equation \ref{i2v} is solved by the Adam optimizer, which can be affected by the step size $\alpha$ and iteration number $I$. Figure \ref{fig_abl_step_iter} shows the results of using Block-2 of Resnet-101 as the perturbed layer of the image model with different step sizes and iteration numbers.
It can be seen that smaller $\alpha$ and $I$ have poorer AASR because of under-fitting. While larger $\alpha$ can achieve better AASR with a smaller $I$. To achieve the best performance, we adopt $\alpha=0.005$ and $I=60$ in subsequent experiments.

\begin{figure}[t]
\centering
    \includegraphics[width=0.75\columnwidth]{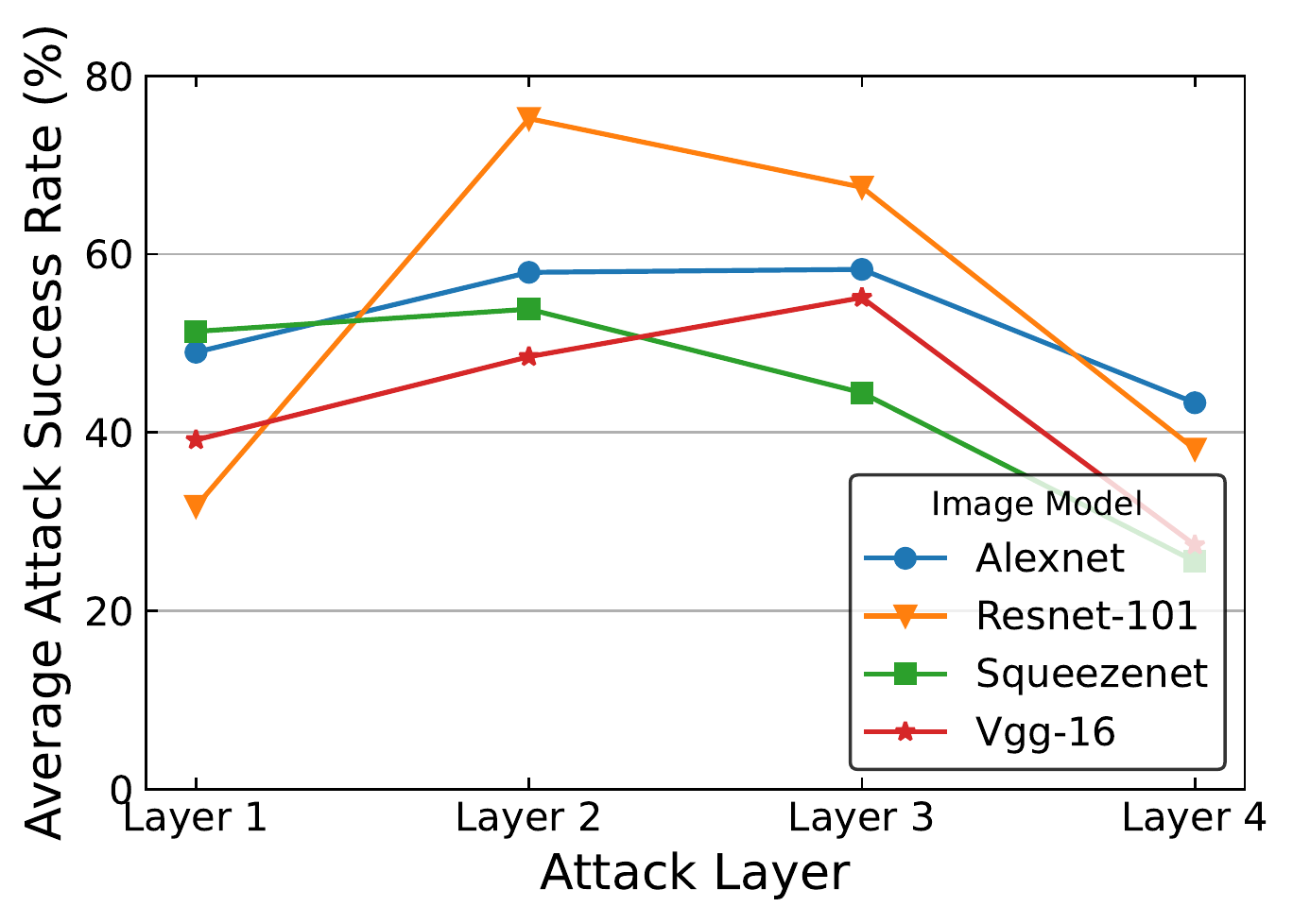}
\caption{AASR (\%) of the I2V attack with various attack layers $l$ in different image models.}
\label{fig_abl_attack_layer}
\end{figure}

\textbf{Intermediate layer selection.} 
For each image model, we select four layers from bottom to top layer (as shown in Table \ref{tab:attacked_layer}) to craft the adversarial perturbations. Figure \ref{fig_abl_attack_layer} shows the results of performing attack on different layers. Attacking the middle layers (layer 2 or layer 3) of image models is better than attacking bottom or top layers. Based on the results, we attack the middle layers, which are marked in \textcolor{red}{Red} in Table \ref{tab:attacked_layer} for each model.

\begin{table}[]
    \centering
    \begin{small}
    \begin{tabular}{l c c c c} \toprule
           Model & Layer 1 & Layer 2 & Layer 3 & Layer 4 \\ \midrule
          Alexnet & ReLU-1 & ReLU-2 & \textcolor{red}{ReLU-3} & ReLU-5 \\ 
          Resnet-101 &  Block-1 & \textcolor{red}{Block-2} & Block-3 & Block-4\\ 
          Squeezenet & Fire-1 & \textcolor{red}{Fire-3} & Fire-5 & Fire-8  \\ 
          Vgg-16 & ReLU-1 & ReLU-5 & \textcolor{red}{ReLU-9} & ReLU-13 \\ 
          \bottomrule
    \end{tabular}
    \caption{Intermediate layer selection. For each image model, the selected intermediate layer used for crafting adversarial perturbations in the proposed I2V attack is in \textcolor{red}{Red}.} 
    \label{tab:attacked_layer}
    \end{small}
\end{table}

\begin{table*}[]
    \centering
    \begin{tabular}{l c c c c c c l}
        \toprule
        \multirow{2}{*}{Attack} & \multirow{2}{*}{Image Model} & \multicolumn{6}{c}{Black-box Video Model} \\ \cmidrule(lr){3-8}
        & & NL-101 & NL-50 & SlowFast-101 & SlowFast-50 & TPN-101 & TPN-50  \\ \midrule
        \multirow{4}{*}{DR \cite{lu2020enhancing}} & Alexnet & 29.70 & 26.73 & 19.80 & 25.74 & 8.91 &	13.86 \\ 
         & Resnet-101 & 14.85	& 23.76	& 18.81	& 27.72	& 15.84	& 20.79 \\ 
         & Squeezenet & 12.87 & 24.75	& 12.87	& 15.84	& 4.95	& 13.86 \\ 
         & Vgg-16 & 14.85	& 29.70	& 13.86	& 22.77	& 6.93	& 15.84\\ \midrule
        \multirow{4}{*}{I2V} & Alexnet & 50.49 & 53.46 & 35.64 & 43.56 & 28.71 & 44.50 \\ 
         & Resnet-101 & \textbf{71.28} & 60.39 & 50.49 & 57.42 & \textbf{61.38} & 71.28 \\
         & Squeezenet & 43.56 & 54.45 & 36.63 & 37.62 & 23.76 & 35.64\\ 
         & Vgg-16 & 43.56 & 48.51 & 28.71 & 39.60 & 21.78 & 32.67	\\ \midrule
        ENS-I2V & Ensemble & \textbf{71.28}	& \textbf{76.23} & \textbf{56.43} & \textbf{62.37} & 52.47	& \textbf{75.24}\\ \bottomrule
    \end{tabular}
    \caption{ASR (\%) against video recognition models on UCF-101.}
    \label{tab:ucf_dr_i2v}
\end{table*}

\begin{table*}[]
    \centering
    \begin{tabular}{l c c c c c c l}
        \toprule
        \multirow{2}{*}{Attack} & \multirow{2}{*}{Image Model} & \multicolumn{6}{c}{Black-box Video Model} \\ \cmidrule(lr){3-8}
        & & NL-101 & NL-50 & SlowFast-101 & SlowFast-50 & TPN-101 & TPN-50  \\ \midrule
        \multirow{4}{*}{DR \cite{lu2020enhancing}} & Alexnet & 22.00 & 31.50 & 43.00 & 41.75 & 31.00 & 39.00 \\ 
         & Resnet-101 & 25.50	& 37.25	& 49.00	& 52.25	& 41.50	& 42.75	 \\ 
         & Squeezenet & 17.00	& 25.00	& 37.00	& 36.50	& 24.25	& 29.50 \\ 
         & Vgg-16 & 16.75	& 23.00	& 36.75	& 35.75	& 23.75	& 29.00	\\ \midrule
        \multirow{4}{*}{I2V} & Alexnet & 44.00 & 54.75 & 61.50 & 59.50 & 59.75 & 69.50 \\
         & Resnet-101 & 56.25 & 64.50 & 74.75 & \textbf{77.00} & \textbf{87.25} & \textbf{90.25} \\ 
         & Squeezenet & 37.75 & 51.00 & 62.50 & 60.25 & 55.50 & 58.50 \\ 
         & Vgg-16 & 39.00 & 46.25 & 57.75 & 59.00 & 59.00 & 70.50	\\ \midrule
        ENS-I2V & Ensemble & \textbf{65.00} & \textbf{72.25} & \textbf{79.75} & 76.50 & 85.75 & 88.00\\ \bottomrule
    \end{tabular}
    \caption{ASR (\%) against video recognition models on Kinetics-400.}
    \label{tab:kin_dr_i2v}
\end{table*}

\subsection{Performance comparison}
Since transferability between hetero-modal models has never been explored, we compare our proposed I2V and ENS-I2V attacks with DR \cite{lu2020enhancing}, which is originally proposed to enhance cross-task transferability. DR minimizes the standard deviation of intermediate features to degrade the recognizability of images. We extend DR to optimize adversarial examples on the Imagenet-pretrained image models and use the same setting as the I2V attack.

The results of attacking UCF-101 and Kinetics-400 datasets are shown in Table \ref{tab:ucf_dr_i2v} and \ref{tab:kin_dr_i2v}, respectively.
From the results, we have the following observations.
First, the proposed I2V and ENS-I2V attacks achieve much higher ASR than DR by a large margin. For example, compared with DR, I2V can boost AASR of more than 63.33\%  and 42.51\% for UCF-101 and Kinetics-400 separately.
Second, I2V using Resnet-101 as the white-box image model outperforms all the other I2V attacks, which suggests that 2D Resnet101 and 3D Resnet-101 in the backbone of video models share more similar feature space than other 2D image models. 
Third, ENS-I2V further improves the average AASR to 65.68\% against UCF101 and 77.88\% against Kinetics-400. Which demonstrates the validity of attacking an ensemble of image models. 
In general, our method, which considers minimizing the cosine similarity between features from adversarial and benign examples, consistently outperforms DR. These experiments validate the effectiveness of the proposed I2V attack.

\subsection{Comparing against stronger baselines}
We further compare the proposed I2V attack against several existing transfer-based attacks that are designed for homomodal models (e.g., images models or video models). It's worthwhile to mention that the comparisons are unfair because the existing transfer-based attacks require the white-box video recognition models to generate the adversarial perturbations. For the comparison, several transfer-based attacks, such as FGSM \cite{goodfellow2014explaining}, BIM \cite{kurakin2016adversarial}, MI \cite{dong2018boosting}, DI \cite{xie2019improving}, TI \cite{dong2019evading}, SIM \cite{lin2019nesterov}, SGM \cite{wu2020skip}, TAP \cite{zhou2018transferable}, ATA \cite{wu2020boosting}, and TT \cite{wei2021boosting} are used as baselines. For these baselines, NL-101, SlowFast-101 and TPN-101 are used as the white-box models. It has been illustrated in ILA \cite{huang2019enhancing}, the transferability of generated adversarial examples can be further enhanced through the proposed fine-tuning methods ILAP and ILAF. Compared to ILAP, ILAF achieves better performance by maintaining the existing adversarial direction and increases the magnitude of feature perturbations under $L_2$ norm \cite{huang2019enhancing}. Therefore, for the compared baseline methods, we use ILAF to fine-tune the generated adversarial examples.

\begin{figure*}
  \centering
  \begin{subfigure}{.3\textwidth}
    \includegraphics[width=1.\linewidth]{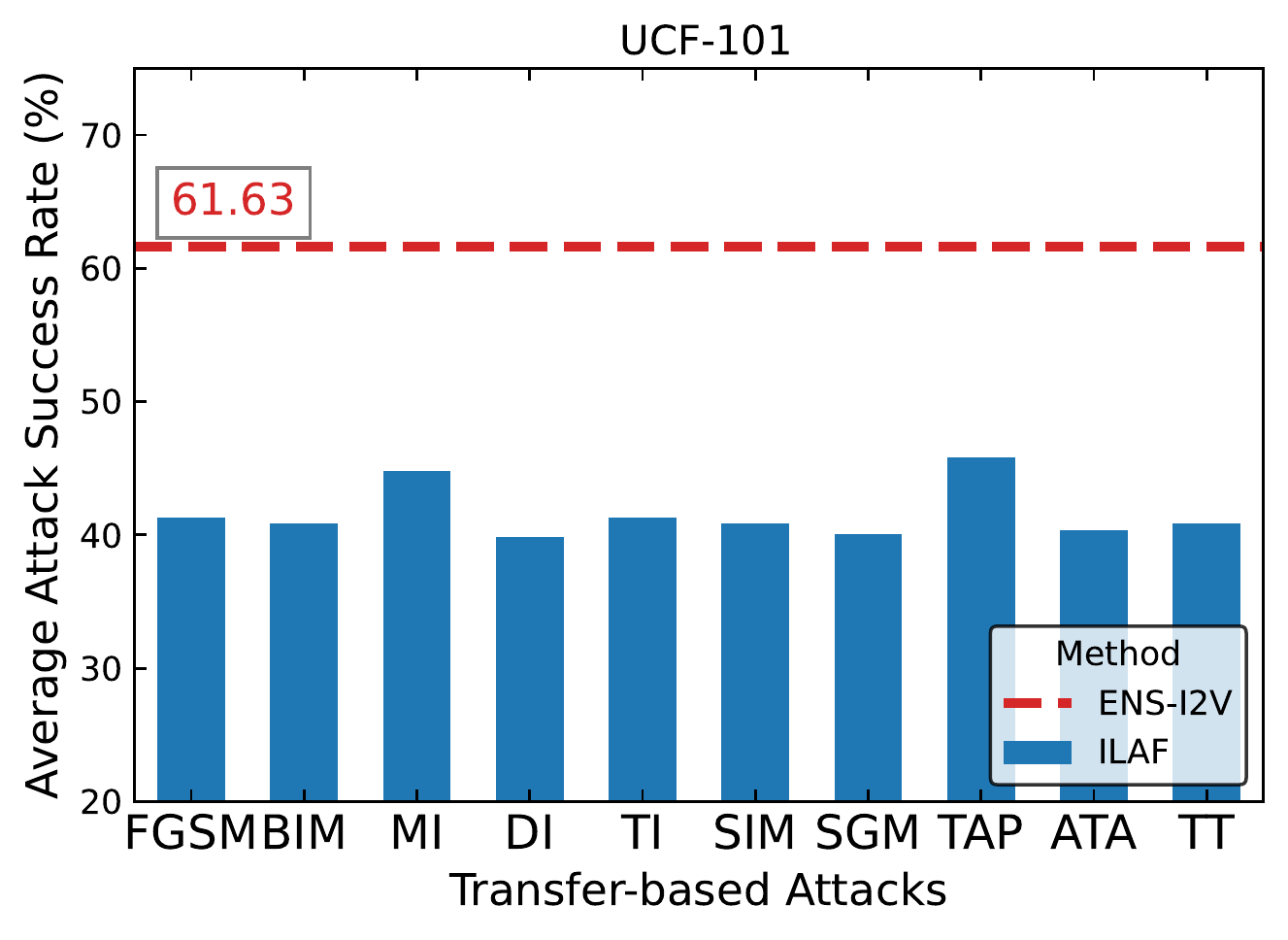}
    \caption{NL-101}
    \label{fig_nl_ucf}
  \end{subfigure}
  \begin{subfigure}{.3\textwidth}
  \includegraphics[width=1.\linewidth]{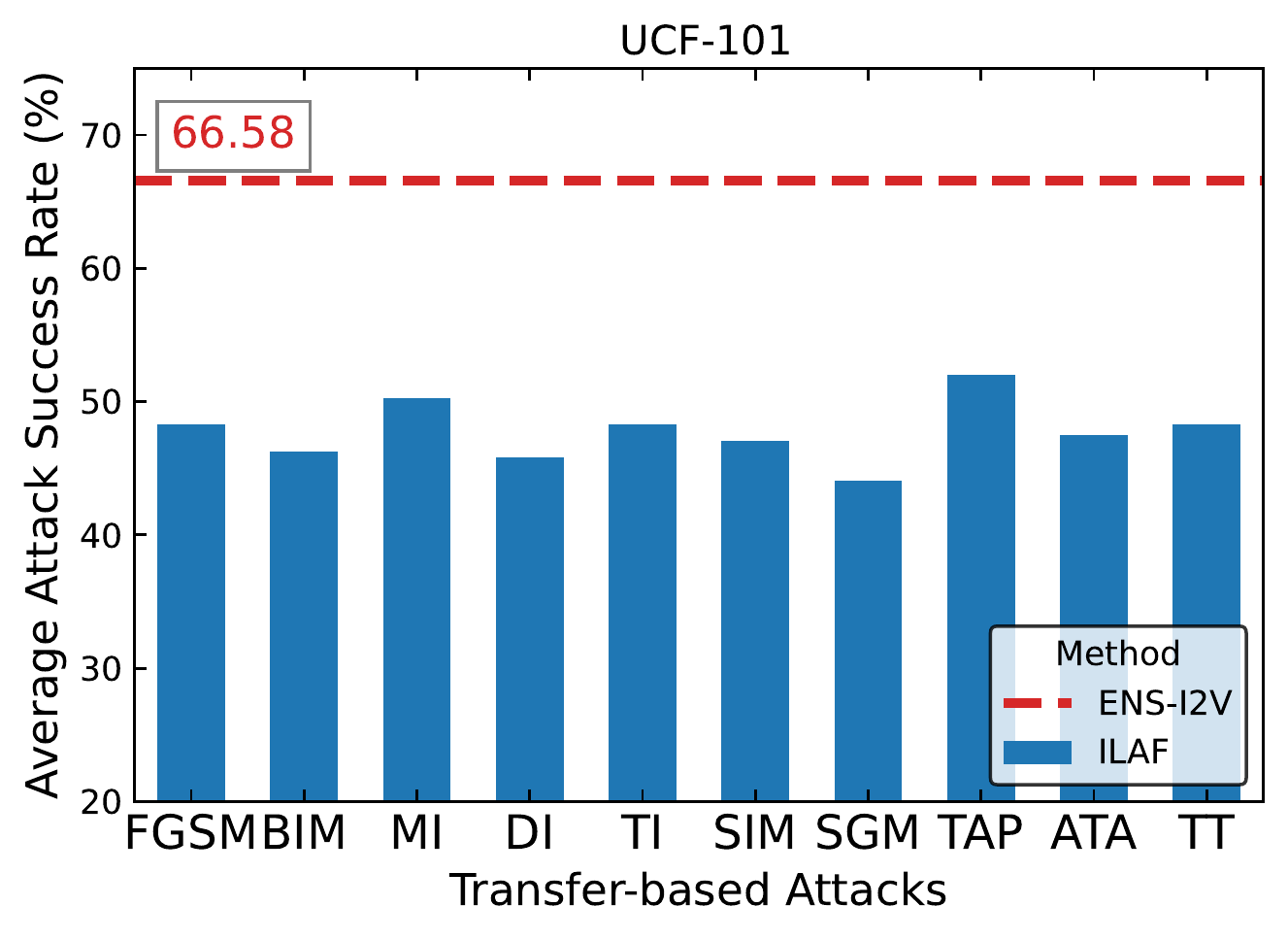}
    \caption{SlowFast-101}
    \label{fig_sf_ucf}
  \end{subfigure}
  \begin{subfigure}{.3\textwidth}
  \includegraphics[width=1.\linewidth]{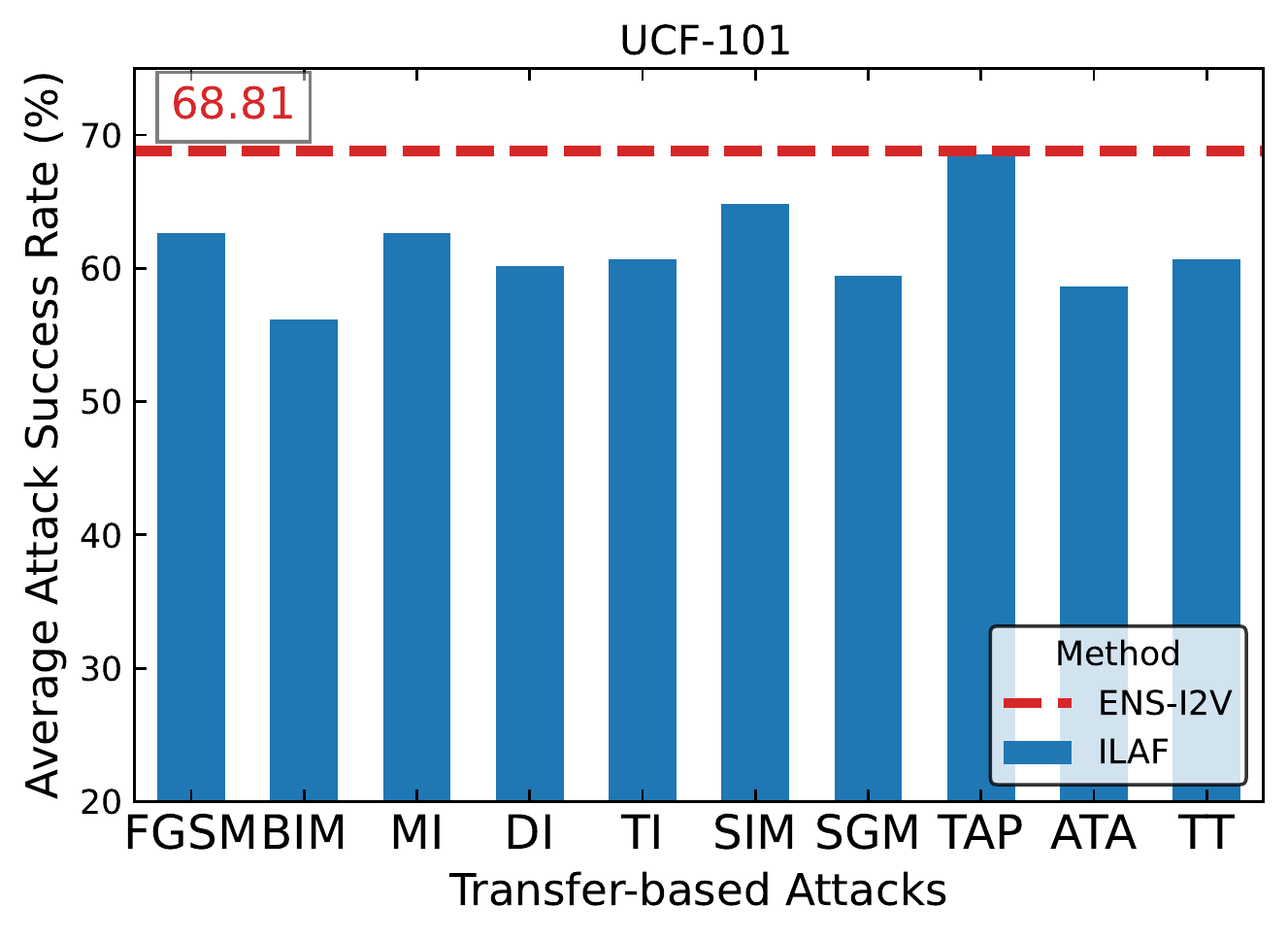}
    \caption{TPN-101}
    \label{fig_tpn_ucf}
  \end{subfigure}
  \begin{subfigure}{.3\textwidth}
  \includegraphics[width=1.\linewidth]{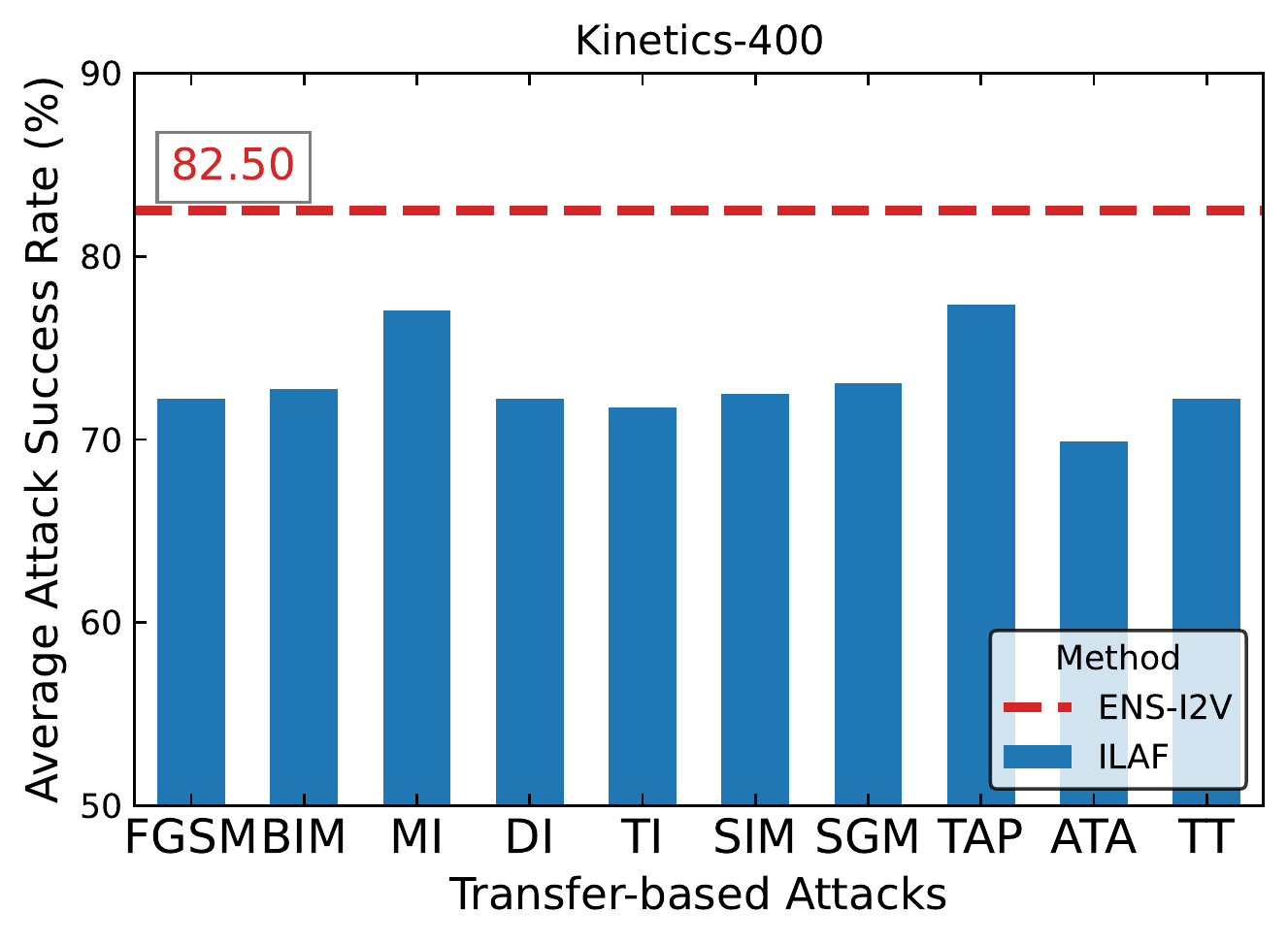}
    \caption{NL-101}
    \label{fig_nl_kin}
  \end{subfigure}
  \begin{subfigure}{.3\textwidth}
  \includegraphics[width=1.\linewidth]{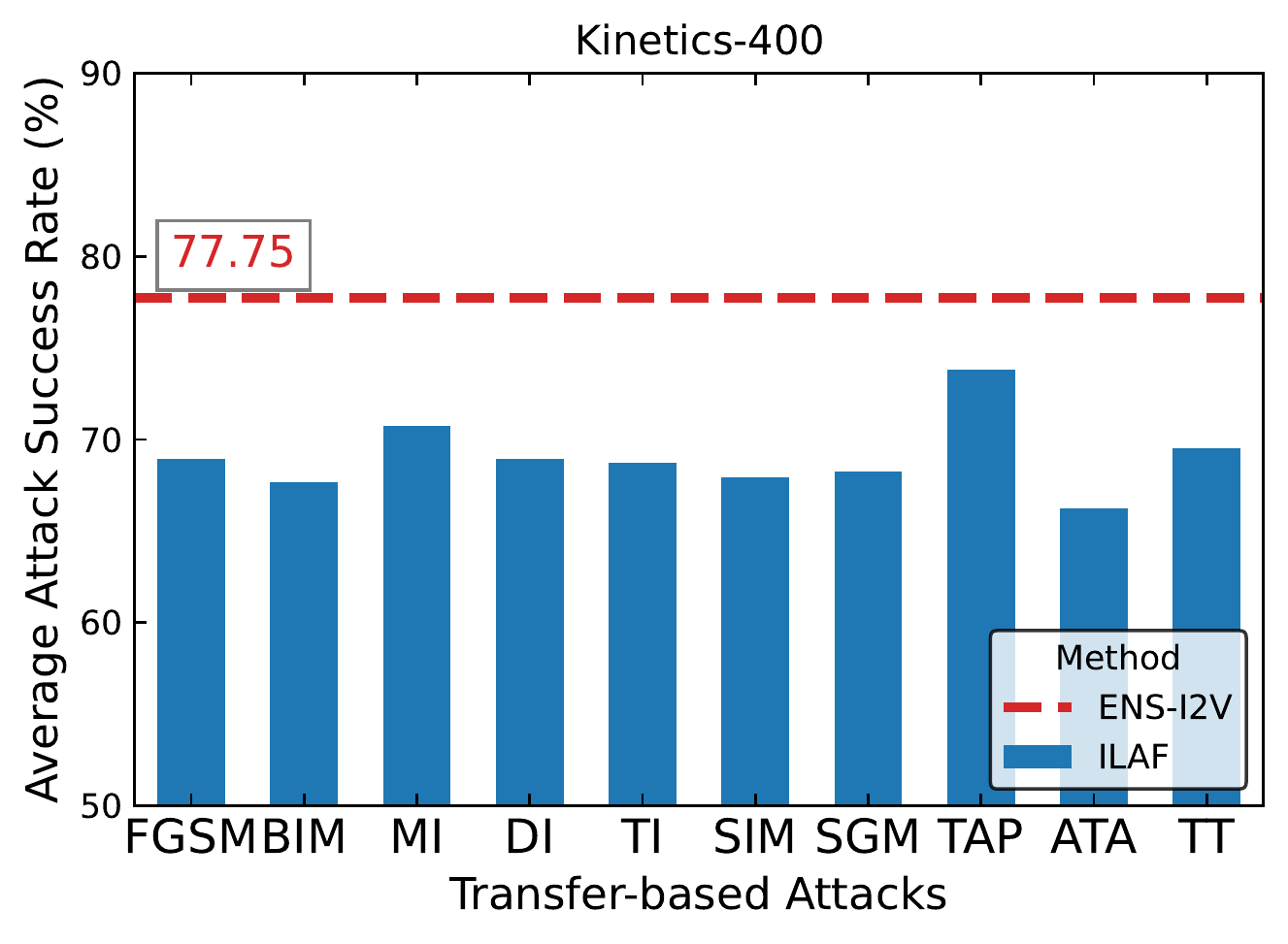}
    \caption{SlowFast-101}
    \label{fig_sf_kin}
  \end{subfigure}
  \begin{subfigure}{.3\textwidth}
  \includegraphics[width=1.\linewidth]{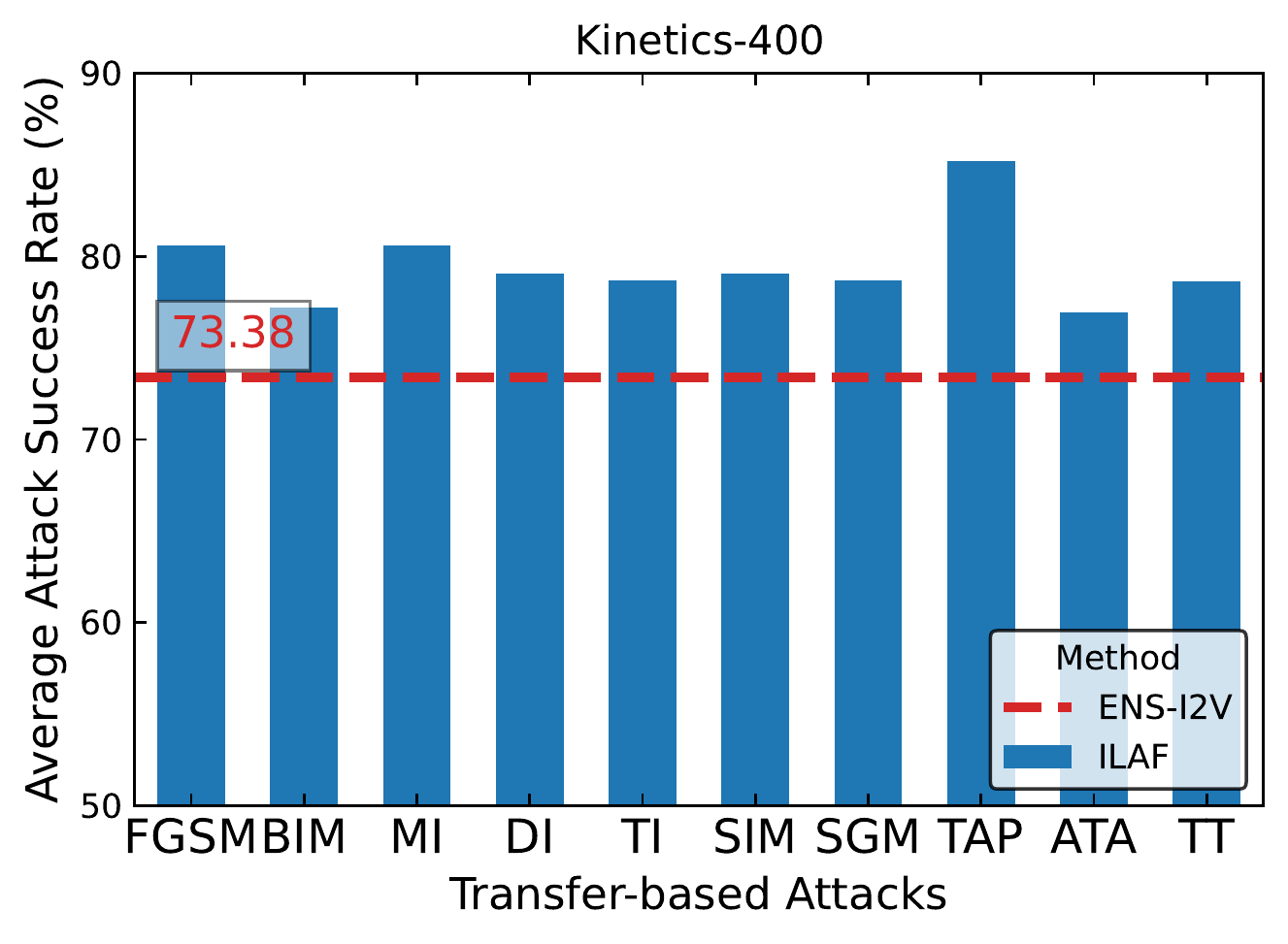}
    \caption{TPN-101}
    \label{fig_tpn_kin}
  \end{subfigure}
  \caption{AASR (\%) against video recognition models with fine-tuning attack methods. 
  The top and bottom rows are the results on UCF-101 and Kinetics-400 respectively. 
  The three columns use the NL-101, SlowFast-101, and TPN-101 models as white-box models respectively. AASR is calculated by averaging ASR over black-box video models that have a different architecture from the white-box model. Red dashed lines denote the performance of our proposed ENS-I2V.}
  \label{fig_strong}
\end{figure*}

Figure~\ref{fig_strong} shows the comparison results. From the results, we have the following observations. First, despite the comparisons being unfair since our method does not require any white-box video models, the proposed ENS-I2V still performs much better than ILAF for most cases. As shown in Figure~\ref{fig_strong}(a)(b)(d)(f), on both UCF-101 and Kinetics-400, ENS-I2V exceeds ILAF by a large margin when using NL-101 or SlowFast-101 as the white-box model. Second, our I2V attack performs worse than baseline methods when they use TPN-101 as the white-box model on Kinetics-400 (Figure ~\ref{fig_tpn_kin}).
This may be because that Kinetics-400 contains richer motion information than UCF-101 and such motion information are unlikely to be well captured by image models. On the contrary, by fusing multi-layer features, TPN-101 can better capture the motion information. As a result, disrupting motion information (Figure~\ref{fig_tpn_kin}) can achieve better performances (see more discussion in Appendix).

\subsection{Discussion}
\begin{figure}
  \centering
  \begin{subfigure}{.22\textwidth}
    \includegraphics[width=1.\linewidth]{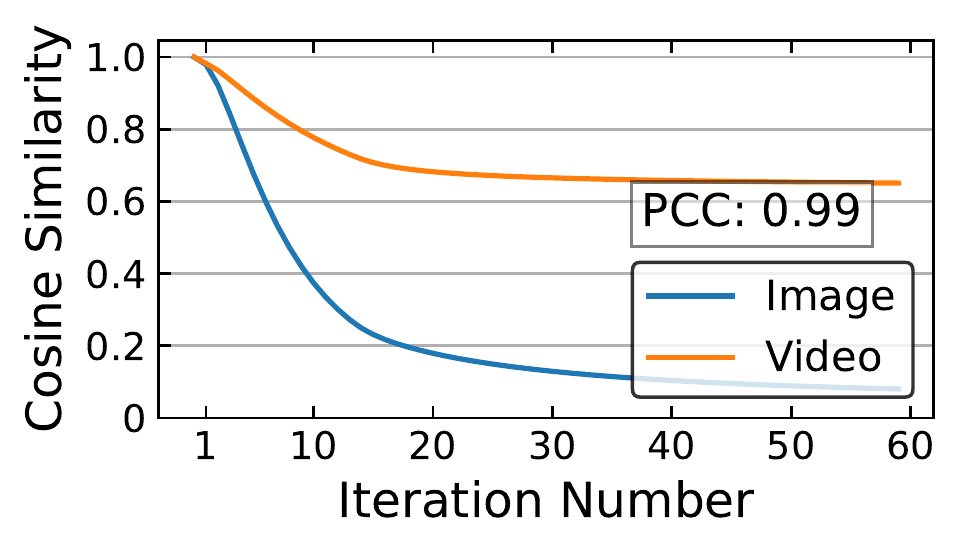}
    \caption{Alexnet}
    \label{fig_alexnet}
  \end{subfigure}
  \begin{subfigure}{.22\textwidth}
  \includegraphics[width=1.\linewidth]{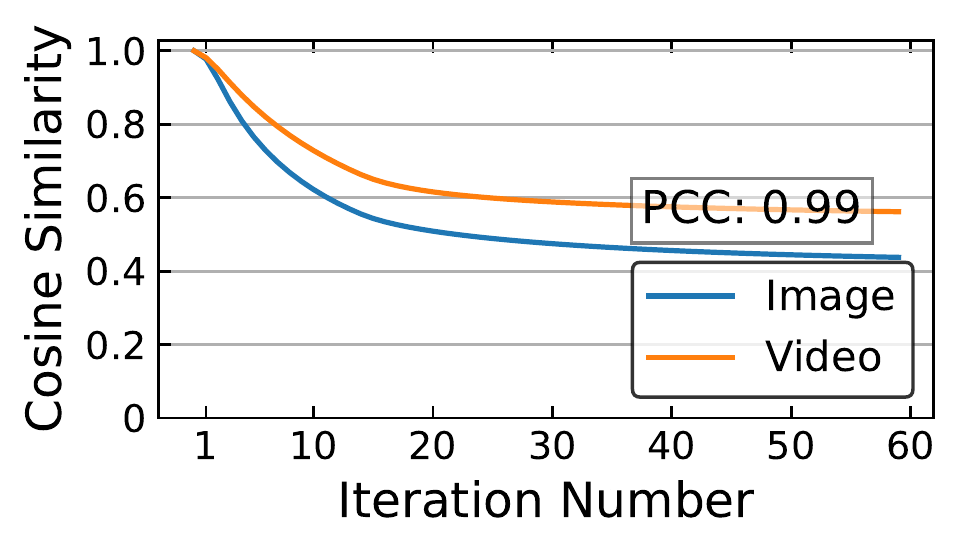}
    \caption{Resnet-101}
    \label{fig_resnet}
  \end{subfigure}
  \begin{subfigure}{.22\textwidth}
  \includegraphics[width=1.\linewidth]{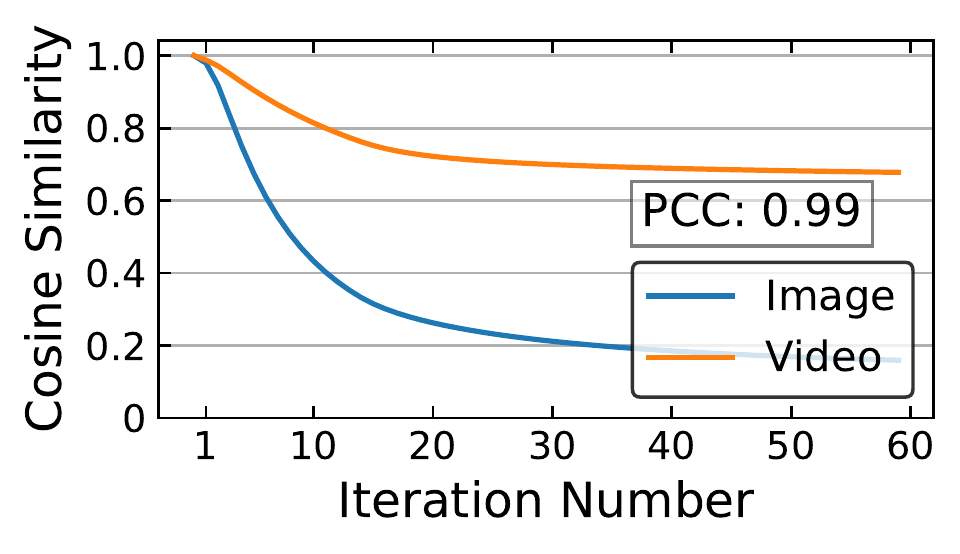}
    \caption{Squeezenet}
    \label{fig_squeezenet}
  \end{subfigure}
  \begin{subfigure}{.22\textwidth}
  \includegraphics[width=1.\linewidth]{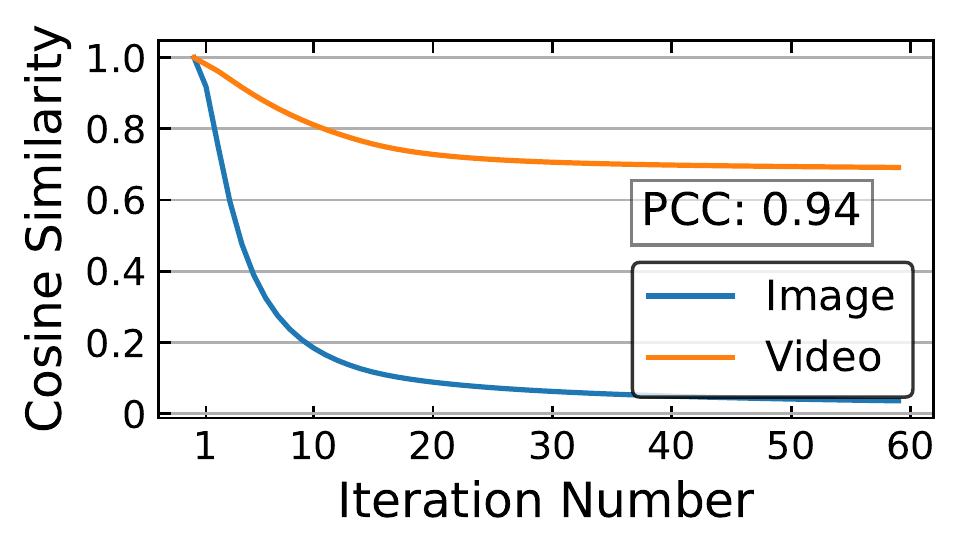}
    \caption{Vgg-16}
    \label{fig_vgg}
  \end{subfigure}
  \caption{
  Pearson correlation coefficient (PCC) analysis between cosine similarity trends computed from image and video models. The used intermediate layers of image models are summarized in Tab.~\ref{tab:attacked_layer} (marked in red). NL-101 uses the second 3D-Resnet block.}
  \label{fig_cs}
\end{figure}

To experimentally demonstrate the effectiveness of the optimized object function (Equation \ref{i2v}), we investigate the changes in the cosine similarity of the adversarial image/video features to the benign image/video features by increasing the iteration number. Pearson Correlation Coefficient (PCC) \cite{anderson1962introduction} is used to measure the linear correlation of cosine similarity trends computed from image and video models.
Figure \ref{fig_cs} shows the PCC analysis of cosine similarity trends using 4 image models and a video model (NL-101).
As can be seen, all PCCs are close to 1, which implies an exact positive linear relationship between the directional changes of image and video intermediate features.
It suggests that minimizing the cosine similarity of image models can make intermediate features of video adversarial examples generated from Imagenet-pretrained models orthogonal to their benign video features. Similar trends are observed when using other video models (see Appendix).

\section{Conclusion}
In this paper, we identify the existence of a similar feature space between image and video models, which can be leveraged to generate adversarial examples from image models to attack black-box video models. 
More specifically, we proposed the Image To Video (I2V) attack, which optimizes adversarial frames on the ImageNet-pretrained image model by minimizing the cosine similarity between features from adversarial and benign examples for perturbing intermediate feature space. Besides, we proposed ENS-I2V to attack an ensemble of image models for boosting transferability. To validate the effectiveness of the proposed methods, we conducted a series of experiments to show that the proposed methods can significantly improve transferability. The results indicate that cross-modal adversarial transferability occurs even across image and video domains.
In the future, we will combine temporal information of videos into image models to further boost transferability.

{\small
\bibliographystyle{ieee_fullname}
\bibliography{egbib}

\begin{thebibliography}{10}\itemsep=-1pt

\bibitem{anderson1962introduction}
Theodore~Wilbur Anderson.
\newblock An introduction to multivariate statistical analysis.
\newblock Technical report, Wiley New York, 1962.

\bibitem{carreira2017quo}
Joao Carreira and Andrew Zisserman.
\newblock Quo vadis, action recognition? a new model and the kinetics dataset.
\newblock In {\em proceedings of the IEEE Conference on Computer Vision and
  Pattern Recognition}, pages 6299--6308, 2017.

\bibitem{dong2018boosting}
Yinpeng Dong, Fangzhou Liao, Tianyu Pang, Hang Su, Jun Zhu, Xiaolin Hu, and
  Jianguo Li.
\newblock Boosting adversarial attacks with momentum.
\newblock In {\em Proceedings of the IEEE conference on computer vision and
  pattern recognition}, pages 9185--9193, 2018.

\bibitem{dong2019evading}
Yinpeng Dong, Tianyu Pang, Hang Su, and Jun Zhu.
\newblock Evading defenses to transferable adversarial examples by
  translation-invariant attacks.
\newblock In {\em Proceedings of the IEEE/CVF Conference on Computer Vision and
  Pattern Recognition}, pages 4312--4321, 2019.

\bibitem{eykholt2018robust}
Kevin Eykholt, Ivan Evtimov, Earlence Fernandes, Bo Li, Amir Rahmati, Chaowei
  Xiao, Atul Prakash, Tadayoshi Kohno, and Dawn Song.
\newblock Robust physical-world attacks on deep learning visual classification.
\newblock In {\em Proceedings of the IEEE conference on computer vision and
  pattern recognition}, pages 1625--1634, 2018.

\bibitem{feichtenhofer2019slowfast}
Christoph Feichtenhofer, Haoqi Fan, Jitendra Malik, and Kaiming He.
\newblock Slowfast networks for video recognition.
\newblock In {\em Proceedings of the IEEE/CVF international conference on
  computer vision}, pages 6202--6211, 2019.

\bibitem{goodfellow2014explaining}
Ian~J Goodfellow, Jonathon Shlens, and Christian Szegedy.
\newblock Explaining and harnessing adversarial examples.
\newblock {\em arXiv preprint arXiv:1412.6572}, 2014.

\bibitem{he2016deep}
Kaiming He, Xiangyu Zhang, Shaoqing Ren, and Jian Sun.
\newblock Deep residual learning for image recognition.
\newblock In {\em Proceedings of the IEEE conference on computer vision and
  pattern recognition}, pages 770--778, 2016.

\bibitem{huang2019enhancing}
Qian Huang, Isay Katsman, Horace He, Zeqi Gu, Serge Belongie, and Ser-Nam Lim.
\newblock Enhancing adversarial example transferability with an intermediate
  level attack.
\newblock In {\em Proceedings of the IEEE/CVF International Conference on
  Computer Vision}, pages 4733--4742, 2019.

\bibitem{iandola2016squeezenet}
Forrest~N Iandola, Song Han, Matthew~W Moskewicz, Khalid Ashraf, William~J
  Dally, and Kurt Keutzer.
\newblock Squeezenet: Alexnet-level accuracy with 50x fewer parameters and< 0.5
  mb model size.
\newblock {\em arXiv preprint arXiv:1602.07360}, 2016.

\bibitem{jiang2019black}
Linxi Jiang, Xingjun Ma, Shaoxiang Chen, James Bailey, and Yu-Gang Jiang.
\newblock Black-box adversarial attacks on video recognition models.
\newblock In {\em Proceedings of the 27th ACM International Conference on
  Multimedia}, pages 864--872, 2019.

\bibitem{jiang2017exploiting}
Yu-Gang Jiang, Zuxuan Wu, Jun Wang, Xiangyang Xue, and Shih-Fu Chang.
\newblock Exploiting feature and class relationships in video categorization
  with regularized deep neural networks.
\newblock {\em IEEE transactions on pattern analysis and machine intelligence},
  40(2):352--364, 2017.

\bibitem{kingma2014adam}
Diederik~P Kingma and Jimmy Ba.
\newblock Adam: A method for stochastic optimization.
\newblock {\em arXiv preprint arXiv:1412.6980}, 2014.

\bibitem{krizhevsky2014one}
Alex Krizhevsky.
\newblock One weird trick for parallelizing convolutional neural networks.
\newblock {\em arXiv preprint arXiv:1404.5997}, 2014.

\bibitem{kurakin2016adversarial}
Alexey Kurakin, Ian Goodfellow, Samy Bengio, et~al.
\newblock Adversarial examples in the physical world, 2016.

\bibitem{lin2019nesterov}
Jiadong Lin, Chuanbiao Song, Kun He, Liwei Wang, and John~E Hopcroft.
\newblock Nesterov accelerated gradient and scale invariance for adversarial
  attacks.
\newblock {\em arXiv preprint arXiv:1908.06281}, 2019.

\bibitem{liu2016delving}
Yanpei Liu, Xinyun Chen, Chang Liu, and Dawn Song.
\newblock Delving into transferable adversarial examples and black-box attacks.
\newblock {\em arXiv preprint arXiv:1611.02770}, 2016.

\bibitem{lu2020enhancing}
Yantao Lu, Yunhan Jia, Jianyu Wang, Bai Li, Weiheng Chai, Lawrence Carin, and
  Senem Velipasalar.
\newblock Enhancing cross-task black-box transferability of adversarial
  examples with dispersion reduction.
\newblock In {\em Proceedings of the IEEE/CVF Conference on Computer Vision and
  Pattern Recognition}, pages 940--949, 2020.

\bibitem{ren2015faster}
Shaoqing Ren, Kaiming He, Ross Girshick, and Jian Sun.
\newblock Faster r-cnn: Towards real-time object detection with region proposal
  networks.
\newblock {\em Advances in neural information processing systems}, 28:91--99,
  2015.

\bibitem{sharif2016accessorize}
Mahmood Sharif, Sruti Bhagavatula, Lujo Bauer, and Michael~K Reiter.
\newblock Accessorize to a crime: Real and stealthy attacks on state-of-the-art
  face recognition.
\newblock In {\em Proceedings of the 2016 acm sigsac conference on computer and
  communications security}, pages 1528--1540, 2016.

\bibitem{simonyan2014very}
Karen Simonyan and Andrew Zisserman.
\newblock Very deep convolutional networks for large-scale image recognition.
\newblock {\em arXiv preprint arXiv:1409.1556}, 2014.

\bibitem{soomro2012ucf101}
Khurram Soomro, Amir~Roshan Zamir, and Mubarak Shah.
\newblock Ucf101: A dataset of 101 human actions classes from videos in the
  wild.
\newblock {\em arXiv preprint arXiv:1212.0402}, 2012.

\bibitem{szegedy2013intriguing}
Christian Szegedy, Wojciech Zaremba, Ilya Sutskever, Joan Bruna, Dumitru Erhan,
  Ian Goodfellow, and Rob Fergus.
\newblock Intriguing properties of neural networks.
\newblock {\em arXiv preprint arXiv:1312.6199}, 2013.

\bibitem{wang2018non}
Xiaolong Wang, Ross Girshick, Abhinav Gupta, and Kaiming He.
\newblock Non-local neural networks.
\newblock In {\em Proceedings of the IEEE conference on computer vision and
  pattern recognition}, pages 7794--7803, 2018.

\bibitem{wei2019sparse}
Xingxing Wei, Jun Zhu, Sha Yuan, and Hang Su.
\newblock Sparse adversarial perturbations for videos.
\newblock In {\em Proceedings of the AAAI Conference on Artificial
  Intelligence}, volume~33, pages 8973--8980, 2019.

\bibitem{wei2020heuristic}
Zhipeng Wei, Jingjing Chen, Xingxing Wei, Linxi Jiang, Tat-Seng Chua, Fengfeng
  Zhou, and Yu-Gang Jiang.
\newblock Heuristic black-box adversarial attacks on video recognition models.
\newblock In {\em Proceedings of the AAAI Conference on Artificial
  Intelligence}, volume~34, pages 12338--12345, 2020.

\bibitem{wei2021boosting}
Zhipeng Wei, Jingjing Chen, Zuxuan Wu, and Yu-Gang Jiang.
\newblock Boosting the transferability of video adversarial examples via
  temporal translation, 2021.

\bibitem{wu2020skip}
Dongxian Wu, Yisen Wang, Shu-Tao Xia, James Bailey, and Xingjun Ma.
\newblock Skip connections matter: On the transferability of adversarial
  examples generated with resnets.
\newblock {\em arXiv preprint arXiv:2002.05990}, 2020.

\bibitem{wu2020boosting}
Weibin Wu, Yuxin Su, Xixian Chen, Shenglin Zhao, Irwin King, Michael~R Lyu, and
  Yu-Wing Tai.
\newblock Boosting the transferability of adversarial samples via attention.
\newblock In {\em Proceedings of the IEEE/CVF Conference on Computer Vision and
  Pattern Recognition}, pages 1161--1170, 2020.

\bibitem{xie2019improving}
Cihang Xie, Zhishuai Zhang, Yuyin Zhou, Song Bai, Jianyu Wang, Zhou Ren, and
  Alan~L Yuille.
\newblock Improving transferability of adversarial examples with input
  diversity.
\newblock In {\em Proceedings of the IEEE/CVF Conference on Computer Vision and
  Pattern Recognition}, pages 2730--2739, 2019.

\bibitem{yang2020temporal}
Ceyuan Yang, Yinghao Xu, Jianping Shi, Bo Dai, and Bolei Zhou.
\newblock Temporal pyramid network for action recognition.
\newblock In {\em Proceedings of the IEEE/CVF Conference on Computer Vision and
  Pattern Recognition}, pages 591--600, 2020.

\bibitem{yue2015beyond}
Joe Yue-Hei~Ng, Matthew Hausknecht, Sudheendra Vijayanarasimhan, Oriol Vinyals,
  Rajat Monga, and George Toderici.
\newblock Beyond short snippets: Deep networks for video classification.
\newblock In {\em Proceedings of the IEEE conference on computer vision and
  pattern recognition}, pages 4694--4702, 2015.

\bibitem{zhou2018transferable}
Wen Zhou, Xin Hou, Yongjun Chen, Mengyun Tang, Xiangqi Huang, Xiang Gan, and
  Yong Yang.
\newblock Transferable adversarial perturbations.
\newblock In {\em Proceedings of the European Conference on Computer Vision
  (ECCV)}, pages 452--467, 2018.

\end{thebibliography}
}

\end{document}